\documentclass[letterpaper]{article} % DO NOT CHANGE THIS
\usepackage{aaai2027}  % DO NOT CHANGE THIS
\usepackage[hyphens]{url}  % DO NOT CHANGE THIS
\usepackage{graphicx} % DO NOT CHANGE THIS
\usepackage{natbib}  % DO NOT CHANGE THIS AND DO NOT ADD ANY OPTIONS TO IT
\usepackage{caption} % DO NOT CHANGE THIS AND DO NOT ADD ANY OPTIONS TO IT
\usepackage{amsmath}
\usepackage{amssymb}
\usepackage{booktabs}
\usepackage{multirow}
\usepackage{placeins}
\newcounter{appx}
\renewcommand{\theappx}{\Alph{appx}}
\newcommand{\appsection}[1]{\refstepcounter{appx}\section{\theappx.\quad #1}}

\usepackage[breaklinks=true]{hyperref}
\usepackage{xcolor}
\definecolor{linkblue}{rgb}{0.0,0.35,0.7}
\hypersetup{colorlinks=true,allcolors=linkblue}
\makeatletter
\expandafter\let\csname ver@hyperref.sty\endcsname\relax
\makeatother

\nocopyright

\title{WA-JEPA: Rethinking the Video JEPA Paradigm for\\World-Action Modeling in Autonomous Driving}

\author{
\normalsize
\mbox{Xinlin Wang\textsuperscript{\rm 1}\thinspace$^{*}$},
\mbox{Yujiao Xiang\textsuperscript{\rm 1,2}\thinspace$^{*}$},
\mbox{Yuheng Zhou\textsuperscript{\rm 1,3}\thinspace$^{*}$},
\mbox{Jingqi Wang\textsuperscript{\rm 1}\thinspace$^{*}$},
\mbox{Minqing Huang\textsuperscript{\rm 1}\thinspace$^{*\ddagger}$},
\mbox{Jiajie Huang\textsuperscript{\rm 1,4}},
\mbox{Dongxu Wei\textsuperscript{\rm 1}\thinspace$^{\dagger}$},
\mbox{Tingguang Zhou\textsuperscript{\rm 1}},
\mbox{Xiyang Wang\textsuperscript{\rm 1}},
\mbox{Gong Chen\textsuperscript{\rm 1,5}},
\mbox{Zhi Xu\textsuperscript{\rm 1}},
\mbox{Feiyang Tan\textsuperscript{\rm 1}},
\mbox{Hangning Zhou\textsuperscript{\rm 1}},
\mbox{Mu Yang\textsuperscript{\rm 1}}}

\affiliations{
\vspace{2pt}
$^1$Afari Intelligent Drive ~~~
$^2$University of Electronic Science and Technology of China \\
$^3$Southeast University ~~~
$^4$Beijing University of Posts and Telecommunications ~~~
$^5$Tianjin University \\[3pt]
$^{*}$Equal contribution, listed in no particular order.
$^{\dagger}$Project lead.
$^{\ddagger}$Corresponding author: \url{mqhuang1211@gmail.com}.}
\begin{document}
\maketitle

\begin{figure*}[!t]
    \centering
    \includegraphics[width=\textwidth]{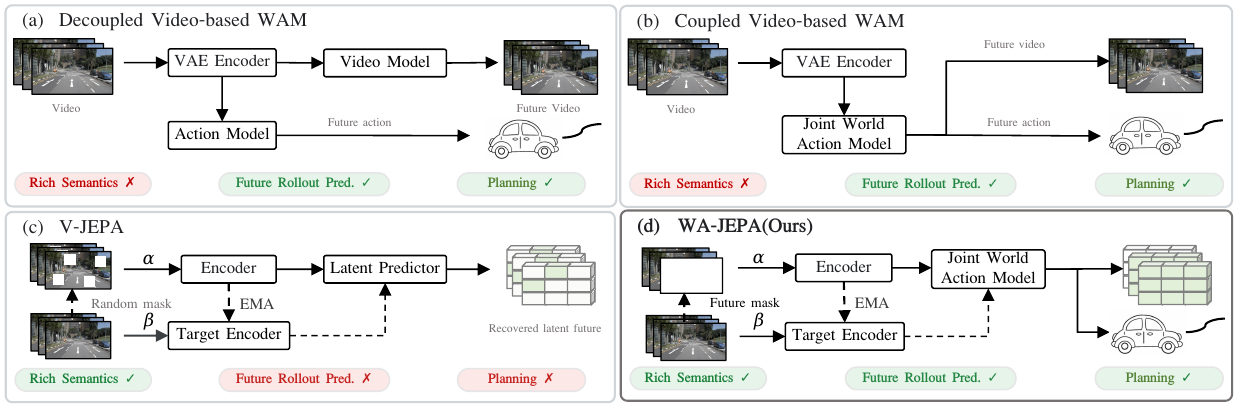}
    \caption{Comparison of V-JEPA, video-based world-action models, and our WA-JEPA. Our method bridges semantic representation learning and predictive world modeling through future-frame masking.}
    \label{fig:method_comparison}
\end{figure*}

\begin{abstract}
Video Joint Embedding Predictive Architecture (V-JEPA) learns powerful spatiotemporal representations from video through self-supervised latent feature prediction. However, V-JEPA is built around random-mask completion and deterministic regression, making it fundamentally ill-suited for autonomous driving planning that demands future-directed prediction tightly coupled with action.
To address this, we rethink the V-JEPA paradigm and present WA-JEPA, a V-JEPA-native world-action model designed for autonomous driving planning.
Instead of random spatiotemporal masking, WA-JEPA employs hybrid future-masked pre-training, where the model infers future latents from observed context.
% Departing from deterministic regression, Moving beyond
% we recast future prediction as flow-based latent diffusion, 
% which substantially improves the model's capability to produce plausible future latents for downstream planning.
Departing from deterministic regression, we recast future prediction as conditional flow matching over latent futures, which substantially improves the model's ability to generate plausible future latents for downstream planning.
Finally, a joint future-action predictor is proposed to denoise future scene tokens and ego trajectories together in a unified spatiotemporal latent space, allowing action supervision to directly shape planning-relevant world representations.
Pre-trained on nuPlan videos and fine-tuned on NAVSIM, WA-JEPA reaches 91.7 EPDMS on NAVSIM-v2, surpassing the strongest end-to-end and world-action baselines by 1.6 and 1.3 EPDMS, and, without HUGSIM-specific fine-tuning, attains the best HD-Score of 0.4462 on the closed-loop HUGSIM benchmark under the same evaluation protocol.
These results validate V-JEPA-native world-action modeling as a powerful and scalable paradigm for autonomous driving planning.
Code is available at \url{https://github.com/AFARI-Research/WA-JEPA}.
\end{abstract}

\section{Introduction}

End-to-end (E2E) autonomous driving distinguishes itself from conventional perception-planning pipelines by learning a unified model that directly maps raw sensor observations to driving actions, thereby eliminating the compounding errors and computational inefficiencies inherent in modular architectures~\cite{hu2023planning}. Despite this conceptual appeal, traditional E2E methods~\cite{jiang2023vad, sun2025sparsedrive} heavily rely on domain-specific perception supervision (e.g., object detection, semantic segmentation, and occupancy prediction) and lack explicit reasoning capabilities, fundamentally limiting their robustness in rare and long-tail driving scenarios.

Recently, two emerging paradigms have sought to endow E2E driving models with reasoning abilities that more closely resemble human-like driving behavior. The first is the Vision-Language-Action (VLA) paradigm~\cite{li2025recogdrive,xu2025wamdiff,li2025drivevlaw0,jia2026driveworldvla}, which leverages the language understanding capacity of Vision-Language Models (VLMs) to make driving decisions from sparse visual observations. In principle, this enables models to reason about complex traffic situations through natural language. However, the majority of VLA methods~\cite{chen2025drivinggpt,li2025imagidrive} simply learn a direct mapping from dense visual inputs to sparse action outputs, suffering from a severe \textit{supervision deficit}~\cite{li2025drivevlaw0}: the vast information bottleneck between rich perceptual signals and minimal action supervision leaves the model poorly constrained.
%A smaller subset of VLA approaches~\cite{xu2024drivegpt4,luo2025adathinkdrive,zhou2026autovla} perform chain-of-thought reasoning in natural language before outputting actions; yet this strategy critically depends on high-quality linguistic annotations of driving reasoning traces, which are prohibitively expensive and difficult to obtain at scale.

The second paradigm, World-Action Models (WAMs)~\cite{li2025drivevlaw0,xia2025drivelaw,huang2026coworld,wang2026latentwam,shi2026drivewam,liu2026driveva}, takes a fundamentally different approach: instead of relying on language as the medium of reasoning, WAMs harness the visual reasoning capability of video generation models~\cite{wan2025wan,HaCohen2024LTXVideo}, i.e., their ability to predict future scene dynamics at pixel level, to inform and guide action decisions. Because future frame prediction naturally provides dense self-supervision without requiring costly perception annotations, WAMs offer a promising solution to the supervision deficit problem~\cite{li2025drivevlaw0}.
%Critically, this paradigm does not depend on linguistic reasoning, making it arguably more aligned with the visual forsight nature of human driving.
Critically, this paradigm does not require linguistic reasoning, making it arguably more aligned with the visual foresight involved in driving.
However, many video-generation-based WAMs perform world modeling in a latent space compressed by a Variational Autoencoder (VAE).
%However, a limitation persists across most existing WAM methods: the world modeling is performed in a latent space highly compressed by a Variational Autoencoder (VAE). 
This impoverished representation space with limited spatiotemporal semantics may constrain the model's ability to understand and reason about the evolving driving world. A question thus arises: \textit{can we enrich the spatiotemporal semantics of the latent representations that underpin WAMs, thereby endowing them with stronger spatiotemporal reasoning and, ultimately, superior driving performance?}

In parallel, a major branch of world model research, the Video Joint Embedding Predictive Architecture (V-JEPA) family~\cite{bardes2024vjepa,assran2025vjepa2}, has developed feature-level masked modeling for self-supervised learning of spatiotemporal representations from massive video corpora. These representations transfer effectively to a wide range of downstream visual understanding tasks, demonstrating remarkable spatiotemporal semantic richness. However, V-JEPA is architecturally ill-suited for planning tasks for three reasons. First, V-JEPA pre-training applies \textit{random} spatiotemporal masking to videos and trains the model to predict the full sequence of features from partially masked context. This is inherently a \textit{completion} objective, which lacks the future-directed predictive capability that is essential for planning. Second, V-JEPA performs this masked prediction via regression, which, while adequate for filling in missing patches within an otherwise observed temporal context, is insufficient for generating entirely unseen future tokens, a task that inherently requires generative modeling. Third, although V-JEPA can be fine-tuned for action-conditioned future prediction~\cite{assran2025vjepa2}, the gap to actionable planning remains vast: existing approaches require a goal image and rely on Model Predictive Control (MPC) with multi-round optimization to recover actions, falling far short of interactive online planning.

In this paper, we rethink the V-JEPA paradigm for world-action modeling in autonomous driving. Our goal is to preserve and leverage V-JEPA's powerful spatiotemporal representation capacity, while fundamentally extending it with causal future generation abilities, and jointly modeling it with action. This yields a novel, V-JEPA-native World-Action Model paradigm, which we term \textbf{WA-JEPA}. Our core technical innovations consist of three key designs:
% \textbf{(1) Hybrid Future-Masked Pre-training.} 
\textbf{(1) Pre-training with Hybrid Future Masking.}
In contrast to the original V-JEPA, which employs random spatiotemporal masking to train a context-completion model, we introduce a carefully designed hybrid future mask strategy: the model observes past frames and predicts the spatiotemporal features of future frames. This endows V-JEPA with future prediction capability, directly aligning the pre-training objective with the forward-predictive demands of planning.
\textbf{(2) V-JEPA-based World Modeling with Flow Matching.} Rather than regressing masked features from context as in the original V-JEPA, we reformulate future prediction as a flow-based generation process over the spatiotemporal feature space. This generative formulation substantially improves the model's ability to produce plausible futures, providing a richer evidentiary basis for downstream planning.
\textbf{(3) V-JEPA-based Joint World-Action Modeling.} Whereas the original V-JEPA models only the visually-observed world, we extend the architecture to jointly model world states and ego actions in a unified spatiotemporal latent space. This tight coupling between visual representations and action representations enables V-JEPA-native future reasoning and decision-making within a single coherent framework.

Our contributions are summarized as follows:
\begin{itemize}
    \item We propose a V-JEPA-native World-Action Model, WA-JEPA, which harnesses V-JEPA's powerful spatiotemporal representations to strengthen joint world-action modeling, achieving substantial improvements in planning capability for autonomous driving.
    % To the best of our knowledge, WA-JEPA is the first driving architecture that uses a unified JEPA-native predictor to jointly model future world states and ego actions in a diffusive latent space.
    % \item We propose WA-JEPA, a V-JEPA-based world-action model that jointly
    % denoises future semantic scene features and ego trajectories within a
    % unified predictor.
    \item We introduce a suite of architectural innovations---hybrid future
masking for pre-training, latent world modeling via flow matching, and joint world-action modeling---that systematically adapt the V-JEPA architecture for planning, while carefully preserving its core representation learning strengths.
    % \item We combine hybrid future-masked pre-training, generative
    % future-feature modeling, and joint scene-action prediction to adapt
    % V-JEPA for direct driving planning.
    \item Extensive experiments on the open-loop NAVSIM‑v1 and NAVSIM‑v2 benchmarks establish a new state of the art, and zero-shot evaluation on the closed-loop simulator HUGSIM~\cite{zhou2024hugsim} shows that the gain carries over to closed-loop driving.
\end{itemize}

\section{Related Work}

\paragraph{End-to-end driving and world-action models.}
E2E driving maps sensor observations directly to ego
motion~\cite{hu2023planning,jiang2023vad,liao2025diffusiondrive}, while VLA
models incorporate language reasoning or VLM-guided trajectory
refinement~\cite{shao2024lmdrive,hwang2024emma,zhou2026opendrivevla,wang2026chainflowvla}.
WAMs augment direct planning with learned future-scene
prediction~\cite{wang2024drivedreamer,li2025law,zhu2026eotwm,wang2026latentwam,hong2026drivefuture,li2025drivevlaw0}.
As illustrated in Fig.~\ref{fig:method_comparison}(a), existing WAMs
explore different choices of world representation and the coupling
between future-scene prediction and action generation. In particular,
coupled video-based WAMs jointly model future visual content and actions
within a shared generative process, using video-generation priors or
video latents~\cite{liu2026driveva,shi2026drivewam}, as shown in
Fig.~\ref{fig:method_comparison}(b). Although this coupling provides a
direct interface between future-scene generation and planning, these
methods inherit the limitations of video-generative latent spaces:
their representations are primarily optimized for visual generation and
reconstruction, which may limit semantic abstraction and the modeling of
action-relevant future states.

\paragraph{Predictive representations for world-action modeling.}
Recent world-action models explore predictive representations beyond
pixel-level reconstruction. Latent-WAM uses spatially aware latent scene
representations for future world modeling and trajectory planning
~\cite{wang2026latentwam}, while DINO-WM predicts future features from a
pre-trained DINOv2 encoder to support action-conditioned planning
~\cite{zhou2024dino}. As shown in Fig.~\ref{fig:method_comparison}(c),
JEPA-based approaches learn powerful spatiotemporal representations by predicting target
embeddings rather than reconstructing~\cite{bardes2024vjepa,assran2025vjepa2}.
However, V-JEPA primarily provides predictive visual representations and
does not directly specify how these representations should be converted
into future actions or planning decisions. Drive-JEPA attempts to bridge
this gap by introducing a V-JEPA encoder into an end-to-end driving model,
but still relies on a separate downstream trajectory planner
~\cite{wang2026drivejepa}. WA-JEPA instead extends predictive
representation learning to jointly predict future world features and ego
trajectories within a shared predictive representation.

\section{Method}
\label{sec:methodology}
% Figure~\ref{fig:overview} provides an overview of the proposed WA-JEPA framework. We first formulate the problem in Section~3.1. Section~3.2 introduces the future-masked causal pre-training stage, where WA-JEPA learns predictive multi-view scene representations in V-JEPA latent space. Section~3.3 presents the V-JEPA-based diffusive world-action modeling stage, which jointly denoises future scene tokens and action tokens through modality-specific joint attention.
% （参考c-jepa）
\subsection{Preliminaries}
% （感觉jepa要单独放一小节比较好）
Our method is greatly inspired by V-JEPA 2~\cite{assran2025vjepa2}, which learns predictive representations in latent space instead of reconstructing raw pixels. As shown in Fig.~\ref{fig:method_comparison} (c), given a masked observation \(\alpha\) and a target observation \(\beta\), V-JEPA 2 trains a predictor to infer the target latent from the context latent:
\begin{equation}
\min_{\theta,\psi}
\left\|
P_{\psi}(E_{\theta}(\alpha)) - \mathrm{sg}(E_{\bar{\theta}}(\beta))
\right\|_1 ,
\end{equation}
where \(E_{\theta}\) is the online encoder, \(E_{\bar{\theta}}\) is the target encoder, \(P_{\psi}\) is the latent predictor, and \(\mathrm{sg}(\cdot)\) denotes stop-gradient. The target encoder is updated with an exponential moving average (EMA):
\begin{equation}
\bar{\theta} \leftarrow \mu \bar{\theta} + (1-\mu)\theta ,
\end{equation}
where \(\mu\) is the EMA momentum coefficient.
%For multi-view driving observations, camera-view and temporal embeddings are added to visual tokens, forming a unified latent space for predictive world modeling.

\begin{figure*}[t]
    \centering
    \includegraphics[width=\textwidth]{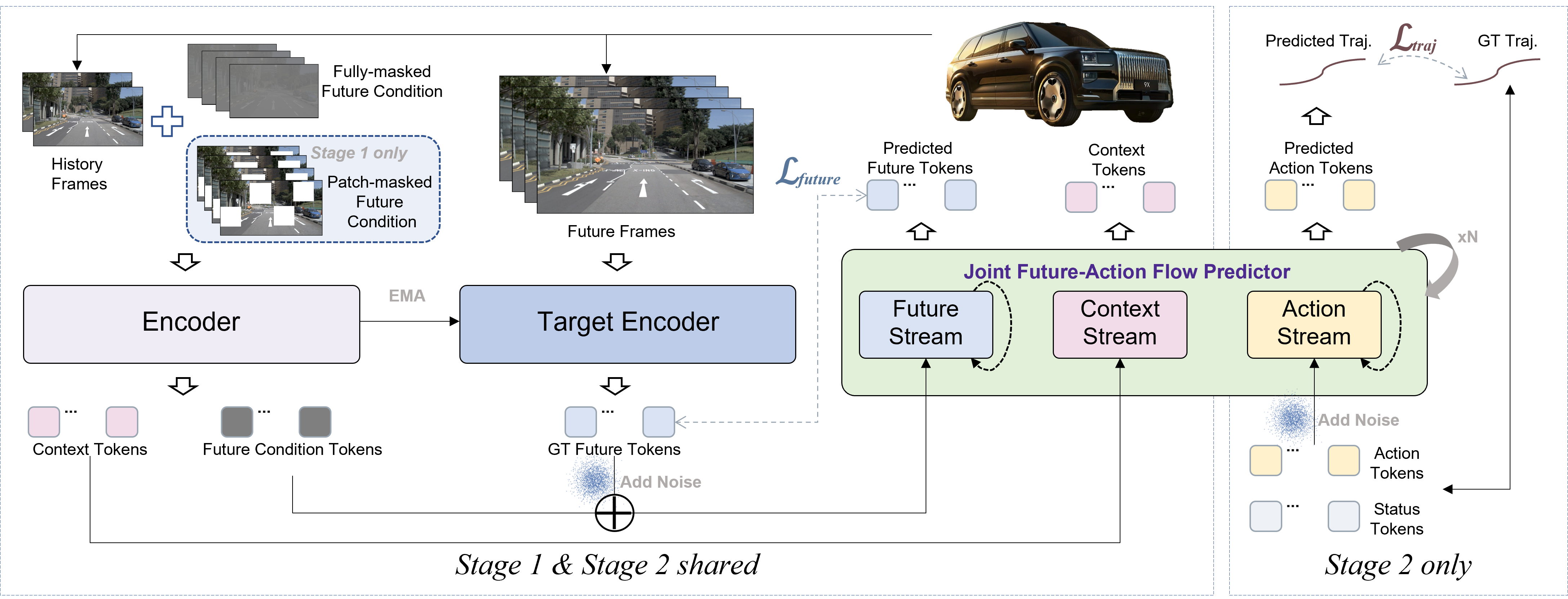}
    % \caption{Overview of WA-JEPA. Stage~1 learns predictive scene
    % representations from multi-view driving videos through future-masked
    % causal diffusion pre-training. Stage~2 initializes from the Stage~1
    % checkpoint and jointly models future scene tokens and ego actions
    % with the Joint Scene-Action Flow Predictor.}
 \caption{Overview of WA-JEPA. Stage~1 adapts V-JEPA 2 to multi-view
    driving videos by predicting future representations under full-future
    and patch-level masks. Stage~2 initializes from this checkpoint and
    jointly predicts future scene representations and ego actions with the
    Joint Future-Action Flow Predictor.}
    \label{fig:framework}
\end{figure*}

\subsection{Overview}
We formulate end-to-end autonomous driving as conditional future-action generation. Given historical multi-view observations \(\mathcal{X}_{1:H}\), ego state \(s\), and historical actions \(\mathcal{Y}_{1:H}\), the model predicts future actions through the following mapping:
% \begin{equation}
% \mathcal{Y}_{1:K}=\{(x_k,y_k,\theta_k)\}_{k=1}^{K},
% \end{equation}
\begin{equation}
f_{\theta}
\left(
\mathcal{X}_{1:H},
s,
\mathcal{Y}_{1:H}
\right)
\mapsto
\hat{\mathcal{Y}}_{H+1:H+K}
=
\left\{
(\hat{x}_k,\hat{y}_k,\hat{\phi}_k)
\right\}_{k=1}^{K},
\end{equation}
where \(f_{\theta}\) denotes the world-action model parameterized by \(\theta\), \(H\) and \(K\) represent the number of historical frames and future frames, respectively. Here, \((\hat{x}_k,\hat{y}_k)\) and $\hat{\phi}_k$ denote the predicted ego position and heading at the \(k\)-th future frame over a 
planning horizon of \(K\) steps. 

% Actions refers to the trajectory in the field of autonomous driving.

% As shown in Fig.~\ref{fig:framework}, the framework consists of two stages. Stage~1, Future-Masked Causal Diffusion Pre-training, learns predictive latent world representations from historical multi-view driving observations by reformulating future latent prediction as a conditional diffusion process that models the distribution of future representations given the observed context. However, these representations alone are insufficient for executable driving actions. Stage~2, Joint World-Action Modeling, bridges this gap by grounding the latent world model with action supervision, jointly modeling future scene tokens and action tokens for end-to-end planning.

As shown in Fig.~\ref{fig:framework}, WA-JEPA follows a two-stage training
scheme. Stage~1 adapts pre-trained V-JEPA 2 to multi-view driving data by
predicting future representations under a hybrid masking strategy.
Stage~2 adds action supervision and extends the predictor with an action
stream, enabling joint prediction of future scene representations and ego
actions.

\subsection{Stage~1: Hybrid Future-Masked Pre-training}

% Stage~1 aims to learn predictive scene representations from synchronized multi-view driving observations without action supervision. Each training sample is formed by historical frames and future frames from the same sequence. The model is trained to predict masked future scene representations in latent space, rather than reconstructing future image pixels. Since the supervision is provided by the feature space of a teacher encoder, this pre-training objective avoids directly modeling low-level appearance details such as color and texture, and instead focuses on future latent scene representation prediction. The parameters learned in Stage~1 are then used to initialize the planning model in Stage~2.
% Stage~1 combines full future masking with patch-masked future completion on synchronized multi-view driving observations without action supervision. The full-mask branch trains strictly past-to-future prediction, whereas the patch-mask branch provides partial future context to facilitate representation learning.

Stage~1 employs a hybrid objective combining a causal Full-mask
branch with a Patch-mask completion branch on synchronized multi-view
driving observations, without action supervision. The former learns
strictly past-to-future dynamics, while the latter retains partial
future context to facilitate representation learning and preserve
the partial-masking paradigm of V-JEPA~2.

\paragraph{Multi-view spatio-temporal encoder.}

We adopt a pre-trained V-JEPA 2 ViT-L as the visual backbone. Each training sample contains historical and future frames from \(C\) synchronized cameras:
\begin{equation}
\mathcal{X}_{1:H+K}
=
\left[
\mathcal{X}_{1:H},
\mathcal{X}_{H+1:H+K}
\right],
\qquad
\mathcal{X}_{t}
=
\{X_t^c\}_{c=1}^{C},
\end{equation}
where \(X_t^c\) represents the frame from the \(c\)-th camera at the \(t\)-th time step.

% Subsequently, the model treats each camera stream as a separate video and processing all views with the shared online encoder for independent encoding and token generation. During training, history tokens remain visible at all times, while masking is applied only to future tokens. Specifically, we employ a hybrid future masking strategy that incorporates both full future masking and patch-masked future masking: the former designates all future tokens as prediction targets, forcing the model to rely solely on historical context for future scenario prediction, while the latter retains a subset of future tokens as visible conditions and predicts the remaining masked tokens, reducing the learning difficulty during early training while following the partial-masking prediction style of V-JEPA.

Subsequently, the model treats each camera stream as a separate video and processes all views independently with the shared online encoder to generate visual tokens. During training, history tokens remain visible, while masking is applied only to future tokens. Specifically, we employ a hybrid strategy that combines full future masking with patch-masked future completion. Full masking designates all future tokens as prediction targets and therefore requires prediction from historical context alone. Patch masking retains a subset of future tokens as visible conditions and predicts the remaining masked tokens, reducing the learning difficulty while retaining the partial-masking prediction style of V-JEPA 2.

The outputs derived from historical frames serve as context scene tokens \(\mathcal{Z}_{\mathrm{ctx}}\) :
\begin{equation}
\mathcal{Z}_{\mathrm{ctx}}
=
E_{\theta}
\left(
\mathcal{X}_{1:H}
\right) .
\end{equation}
% For future frames, given the current future mask pattern \(\mathcal{M}^{(m)}\), where \(m\in\{\mathrm{full},\mathrm{patch}\}\), the model constructs future condition scene tokens \(\mathcal{Z}_{\mathrm{cond}}^{(m)}\) based on the obtained visible future tokens and learnable mask tokens \(\mathcal{Z}_{\mathrm{mask}}\) :

% For future frames, the mask pattern $M^{(m)}$, where
% $m\in\{\mathrm{full},\mathrm{patch}\}$, is applied before the
% online encoder, such that only visible future patches are encoded.
% The resulting visible tokens are combined with learnable mask tokens
% $Z_{\mathrm{mask}}$ to construct the future condition tokens:
% \begin{equation}
% \mathcal{Z}_{\mathrm{cond}}^{(m)}
% =
% \Phi^{(m)}
% \left(
% \mathcal{Z}_{\mathrm{mask}},
% E_{\theta}
% \left(
% \mathcal{X}_{H+1:H+K},
% \mathcal{M}^{(m)}
% \right)
% \right),
% \end{equation}
% where \(\Phi^{(m)}(\cdot)\) denotes a mask-aware fill-and-scatter operation. It first initializes a full future-token sequence with the learnable mask tokens and then scatters the encoded visible future tokens back to their original positions according to the mask pattern. The remaining masked positions retain the learnable mask tokens.

For future frames, the mask pattern $M^{(m)}$, where
$m\in\{\mathrm{full},\mathrm{patch}\}$, is applied before the
online encoder, such that only visible future patches are provided
to $E_\theta$. The model then combines the encoded visible tokens
with learnable mask tokens $Z_{\mathrm{mask}}$ to construct the
future condition tokens:
\begin{equation}
Z_{\mathrm{cond}}^{(m)}
=
\Phi^{(m)}
\left(
Z_{\mathrm{mask}},
E_\theta
\left(
X_{H+1:H+K}, M^{(m)}
\right)
\right),
\end{equation}
where $\Phi^{(m)}(\cdot)$ denotes a mask-aware fill-and-scatter
operation. It initializes a full future-token sequence with learnable
mask tokens and scatters the encoded visible future tokens back to
their original positions according to the mask pattern. Under
Full-mask, no future patch is provided to the online encoder, and
$Z_{\mathrm{cond}}^{(\mathrm{full})}$ therefore consists entirely
of learnable mask tokens.

Meanwhile, the EMA target encoder provides unmasked target representations
for the future frames, yielding the clean future scene target
\(\mathcal{Z}_{\mathrm{future}}^{*}\):

% where \(\Phi(\cdot)\) denotes a mask-aware fill-and-scatter operation. It first initializes a full future-token sequence with the learnable mask tokens, and then scatters the encoded visible future tokens back to their original positions according to the mask pattern. The remaining masked positions retain the learnable mask tokens.

% Meanwhile, the EMA target encoder receives the complete video clip and extracts the tokens corresponding to the future, yielding the clean future scene target \(\mathcal{Z}_{\mathrm{future}}^{*}\) :
\begin{equation}
\mathcal{Z}_{\mathrm{future}}^{*}
=
E_{\bar{\theta}}
\left(
\mathcal{X}_{H+1:H+K}
\right) .
\end{equation}
These target representations is used solely to provide latent supervision for future scenes.

\paragraph{flow-based latent future prediction.}
We use flow matching for future latent prediction.
We use conditional flow matching with a clean-latent
($x$-prediction) parameterization for future latent prediction.
First, we sample Gaussian noise \(\epsilon_{\mathrm{future}} \sim \mathcal{N}(0,I)\) with the same dimensionality as the future scene tokens and continuous flow time \(t\). The noisy future scene tokens \(\mathcal{Z}_{t}\) at time \(t\) is obtained via linear interpolation:
\begin{equation}
\mathcal{Z}_{t}
=
(1-t)\epsilon_{\mathrm{future}}
+
t\mathcal{Z}_{\mathrm{future}}^{*}.
\end{equation}

The future flow predictor receives the historical context \(\mathcal{Z}_{\mathrm{ctx}}\), future condition \(\mathcal{Z}_{\mathrm{cond}}\), noisy future scene tokens \(\mathcal{Z}_{t}\), and temporal condition \(t\), 
% and estimates the denoised future scene tokens \(\hat{\mathcal{Z}}_{\mathrm{future}}\) :
and predicts the clean future scene tokens
$\hat{Z}_{\mathrm{future}}$:
\begin{equation}
\hat{\mathcal{Z}}_{\mathrm{future}}
=
P_{\psi}^{\mathrm{future}}
\left(
\mathcal{Z}_{\mathrm{ctx}},
\mathcal{Z}_{\mathrm{cond}},
\mathcal{Z}_{t},
t
\right).
\end{equation}

The future flow predictor follows an MMDiT (Multimodal Diffusion Transformer)-style~\cite{esser2024scaling} design, primarily performing joint self-attention between context scene tokens and future scene tokens.

\paragraph{Training objective.}
In Stage~1, the model is pre-trained on multi-view driving observations from the nuPlan dataset. 
% During this stage, the model is optimized using an mean squared error (MSE) future latent prediction loss:
During this stage, the model is optimized using a mean squared
error (MSE) loss on the clean future latent prediction:
\begin{equation}
  \mathcal{L}_{\mathrm{Stage~1}}
  =
  \mathcal{L}_{\mathrm{future}}
  =
  \frac{1}{N}
  \left\|
  \hat{\mathcal{Z}}_{\mathrm{future}}
  -
  \mathrm{sg}
  \left(
  \mathcal{Z}_{\mathrm{future}}^{*}
  \right)
  \right\|_2^2,
  \end{equation}
% where \(N\) denotes the number of future scene tokens included in the loss computation.
% After Stage~1 pre-training, the model obtains a predictive multi-view driving scene representation initialization, which is further fine-tuned with action supervision for planning in Stage~2.

where $N$ denotes the number of future scene tokens included in the
loss computation. This clean-latent objective corresponds to the
$x$-prediction parameterization of conditional flow matching
described above. 
% The resulting Stage~1 checkpoint is used to initialize Stage~2,
% where the action stream is introduced and the model is jointly
% fine-tuned with the existing future-latent supervision and additional
% action supervision.

\subsection{Stage~2: Joint World-Action Modeling}

Stage~2 introduces an action generation stream, jointly modeling future scene representations and actions within a unified model. The model is initialized using the pre-trained weights from Stage~1 and further adapted to the action planning task through action supervision.
%(放在overview)

\paragraph{Joint future-action predictor.}
In Stage~2, the joint predictor further incorporates historical actions and the compact ego state. Together, they serve as ego-motion conditions for future actions generation. 
Different from Stage~1, only full future mask is used in Stage~2 for consistency with the causal nature of driving. Future images are used only to construct supervision signals for the future scene latent prediction and are not provided as input to the student predictor.

For the action stream, we first normalize the ground-truth future actions and construct noisy actions in the normalized action space:
\begin{equation}
\begin{gathered}
\tilde{\mathcal{Y}}_{t}
=
(1-t)\epsilon_y
+
t\bar{\mathcal{Y}}_{H+1:H+K},
\quad
\epsilon_y\sim\mathcal{N}(0,I),
\\[3pt]
\bar{\mathcal{Y}}_{H+1:H+K}
=
\mathrm{Norm}(\mathcal{Y}_{H+1:H+K}),
\end{gathered}
\end{equation}
where $\mathrm{Norm}(\cdot)$ denotes the action normalization operation,
$\bar{\mathcal{Y}}_{H+1:H+K}$ denotes the normalized ground-truth future
actions, $\epsilon_y$ denotes Gaussian noise with the same dimensionality
as the future actions, and $\tilde{\mathcal{Y}}_t$ denotes the noisy
future actions at time $t$.

Subsequently, the noisy future actions \(\tilde{\mathcal{Y}}_{t}\), historical actions \(\mathcal{Y}_{1:H}\), and ego state \(s\) are separately encoded and concatenated to form the action tokens $\mathcal{T}_{\mathrm{act}}$:
\begin{equation}
\begin{gathered}
\mathcal{T}_{\mathrm{act}}=
\operatorname{Concat}
\left[
\mathcal{T}_{n},
\mathcal{T}_{h},
\mathcal{T}_{s}
\right],
\\
\mathcal{T}_{n}
=F_{n}(\tilde{\mathcal{Y}}_{t}),\quad
\mathcal{T}_{h}=F_{h}(\mathcal{Y}_{1:H}),\quad
\mathcal{T}_{s}=F_{s}(s).
\end{gathered}
\end{equation}
where \(F_n\) is a linear projection, and \(F_h\) and \(F_s\)
are MLP encoders.

The joint predictor models interactions among the historical context scene tokens \(\mathcal{Z}_{\mathrm{ctx}}\), future scene tokens \(\mathcal{Z}_{\mathrm{cond}}\) and noisy future scene tokens
$\mathcal{Z}_t$, and action tokens \(\mathcal{T}_{\mathrm{act}}\) at time \(t\). Its future scene and action output streams are respectively defined as

\begin{equation}
\begin{gathered}
\hat{\mathcal{Z}}_{\mathrm{future}}
=
P_{\psi}^{\mathrm{future}}
\left(
\mathcal{Z}_{\mathrm{ctx}},
\mathcal{Z}_{\mathrm{cond}},
\mathcal{Z}_t,
\mathrm{sg}\!\left(
\mathcal{T}_{\mathrm{act}}
\right),
t
\right),
\\[3pt]
\hat{\bar{\mathcal{Y}}}_{H+1:H+K}
=
P_{\psi}^{\mathrm{act}}
\left(
\mathcal{Z}_{\mathrm{ctx}},
\mathcal{Z}_{\mathrm{cond}},
\mathcal{Z}_{t},
\mathcal{T}_{\mathrm{act}},
t
\right).
\end{gathered}
\end{equation}

Here, $P_{\psi}^{\mathrm{future}}$ and
$P_{\psi}^{\mathrm{act}}$ denote the future scene and action output streams of the same joint predictor, which also follows an MMDiT-style design, rather than two independent predictors.
Under the Full-mask setting used in Stage~2,
$Z_{\mathrm{cond}}=Z_{\mathrm{cond}}^{(\mathrm{full})}$ consists
entirely of learnable mask tokens, so no future image content is
visible to the joint predictor.
$\hat{\bar{\mathcal{Y}}}_{H+1:H+K}$ denotes the predicted normalized future actions. We apply stop-gradient $\mathrm{sg}(\cdot)$ to the action tokens only in the future scene output stream, preventing the future scene prediction loss from updating
the action stream. Further details and analysis of this design are provided in Appendix~\ref{app:additional-results}.
% ~\ref{app:gradient_design}.
% Although the scene output stream conditions on the action tokens during the forward pass, gradients from the scene prediction loss are blocked at the action-token interface and therefore do not update the action stream. In contrast, the action output stream attends to differentiable historical context and future scene tokens, allowing action supervision to shape the learned scene representations through the joint interaction module. This asymmetric gradient design preserves future scene prediction while encouraging the scene representations to capture future dynamics that are relevant to ego planning.

\paragraph{Training objective.}
The future scene stream follows the future scene prediction loss defined in Stage~1.
This objective preserves the model's future scene modeling capability
during action-supervised fine-tuning. The action stream predicts denoised future actions in the normalized action space and is optimized using the MSE between the predicted and ground-truth actions over all future steps:

\begin{equation}
\begin{gathered}
\mathcal{L}_{\mathrm{act}}
=
\frac{1}{K}
\sum_{k=1}^{K}
\left\|
\hat{\bar{\mathbf{y}}}_{k}
-
\bar{\mathbf{y}}_{k}
\right\|_{2}^{2},
\\[3pt]
\bar{\mathbf{y}}_{k}
=
\mathrm{Norm}
\left(
x_k,
y_k,
\phi_k
\right),
\end{gathered}
\end{equation}
where $\bar{\mathbf{y}}_{k}$ denotes the normalized ground-truth
action at the $k$-th future frame,
$\hat{\bar{\mathbf{y}}}_{k}$ denotes the corresponding predicted
normalized action.

Finally, the overall Stage~2 training objective is
\begin{equation}
\mathcal{L}_{\mathrm{Stage~2}}
=
\lambda_{\mathrm{future}}
\mathcal{L}_{\mathrm{future}}
+
\lambda_{\mathrm{act}}
\mathcal{L}_{\mathrm{act}}.
\end{equation}
Here, $\lambda_{\mathrm{future}}$ and
$\lambda_{\mathrm{act}}$ are the weighting coefficients for the future scene and action losses, respectively.

% \paragraph{Planning inference.}
% During inference, the model takes only historical multi-view observations,
% historical actions, and ego state as inputs. The future scene latents and
% future actions are initialized from Gaussian noise and iteratively denoised by the joint predictor. Neither future images nor ground-truth future actions are required during this process.

% After denoising, the predicted normalized actions are transformed back to
% the original action space to obtain the final future actions:
% \begin{equation}
% \hat{\mathcal{Y}}_{H+1:H+K}
% =
% \mathrm{Norm}^{-1}
% \left(
% \hat{\bar{\mathcal{Y}}}_{H+1:H+K}
% \right).
% \end{equation}
% Here, $\mathrm{Norm}^{-1}(\cdot)$ denotes the action denormalization
% operation. 
\paragraph{Planning inference.}
During inference, the model takes only historical multi-view
observations, historical actions, and ego state as inputs. The future
scene latents and normalized future actions are initialized from
Gaussian noise. At each sampling step, the joint predictor estimates
their clean endpoints, which are converted into velocities along the
linear flow paths and iteratively integrated. Neither future images
nor ground-truth future actions are required during this process.

After sampling, the predicted normalized actions are transformed back
to the original action space to obtain the final future actions:
\begin{equation}
\hat{\mathcal{Y}}_{H+1:H+K}
=
\mathrm{Norm}^{-1}
\left(
\hat{\bar{\mathcal{Y}}}_{H+1:H+K}
\right).
\end{equation}
Here, $\mathrm{Norm}^{-1}(\cdot)$ denotes the action denormalization
operation.

\begin{table*}[t]
    \centering
    {\small
    \setlength{\tabcolsep}{3pt}
    \begin{tabular}{@{}llrrrrrrrrrrr@{}}
        \toprule
        Method & Backbone & NC$\uparrow$ & DAC$\uparrow$ & DDC$\uparrow$ &
        TLC$\uparrow$ & EP$\uparrow$ & TTC$\uparrow$ & LK$\uparrow$ &
        HC$\uparrow$ & EC$\uparrow$ & EPDMS$^{*}\uparrow$ &
        EPDMS$\uparrow$ \\
        \midrule
        \multicolumn{13}{@{}l}{\textbf{End-to-End Methods}} \\
        TransFuser~\cite{chitta2022transfuser} & RegNetY-3.2GF & 96.9 & 89.9 & 97.8 &
        99.7 & 87.1 & 95.4 & 92.7 & 98.3 & 87.2 & 76.7 & -- \\
        ARTEMIS~\cite{feng2025artemis} & ResNet-34 & 98.3 & 95.1 & 98.6 &
        99.8 & 81.5 & 97.4 & 96.5 & 98.3 & -- & 83.1 & -- \\
        Hydra-MDP++~\cite{li2025hydramdpplusplus} & V2-99 & 98.8 & 97.8 &
        99.1 & 100 & 84.0 & 95.3 & 70.1 & -- & 96.8 & 84.1 & -- \\
        DiffusionDrive~\cite{liao2025diffusiondrive} & ResNet-34 & 98.2 & 95.9 &
        99.4 & 99.8 & 87.5 & 97.3 & 96.8 & 98.3 & 87.7 & -- & 84.5 \\
        Drive-JEPA~\cite{wang2026drivejepa} & ResNet-34 & 98.8 & 97.4 & 99.0 &
        99.8 & 83.5 & 98.0 & 96.2 & 98.1 & 85.6 & 85.4 & -- \\
        Drive-JEPA$^\dagger$~\cite{wang2026drivejepa} & ViT-L & 98.4 & 98.6 &
        99.1 & 99.8 & 88.4 & 97.8 & 97.6 & 97.9 & 84.8 & 87.8 & -- \\
        DiffusionDriveV2~\cite{zou2025diffusiondrivev2} & ResNet-34 & 97.7 &
        96.6 & 99.2 & 99.8 & 88.9 & 97.2 & 96.0 & 97.8 & 91.0 &
        85.5 & 87.5 \\
        DriveSuprim~\cite{yao2025drivesuprim} & V2-99 & 97.8 & 97.9 & 99.5 &
        99.9 & 90.6 & 97.1 & 96.6 & 98.3 & 77.9 & 86.0 & -- \\
        SparseDriveV2~\cite{sun2026sparsedrivev2} & ResNet-34 & 98.1 & 98.1 &
        99.6 & 99.8 & 91.1 & 97.3 & 96.9 & 98.2 & 78.4 &
        86.7 & 90.1 \\
        \midrule
        \multicolumn{13}{@{}l}{\textbf{VLA Methods}} \\
        ReCogDrive~\cite{li2025recogdrive} & InternVL3 & 98.3 & 95.2 & 99.5 &
        99.8 & 87.1 & 97.5 & 96.6 & 98.3 & 86.5 & 83.6 & -- \\
        WAM-Flow~\cite{xu2025wamflow} & Janus-1.5B & 98.5 & 94.5 & 99.5 &
        99.8 & 86.9 & 96.8 & 97.4 & 97.6 & 73.9 & 84.7 & -- \\
        WAM-Diff~\cite{xu2025wamdiff} & LLaDA-V & 99.0 & 98.4 & 99.3 &
        99.9 & 87.0 & 98.6 & 96.2 & 98.1 & 78.5 & -- & 89.7 \\
        \midrule
        \multicolumn{13}{@{}l}{\textbf{World-Action and World-Model Methods}} \\
        DriveVLA-W0~\cite{li2025drivevlaw0} & Emu3-8B & 98.5 & 99.1 & 98.0 &
        99.7 & 86.4 & 98.1 & 93.2 & 97.9 & 58.9 & 86.1 & -- \\
        DriveWorld-VLA~\cite{jia2026driveworldvla} & InternVL & 98.6 & 99.1 &
        99.6 & 99.8 & 87.4 & 97.9 & 97.0 & 97.8 & 78.6 & -- & 86.8 \\
        CoWorld-VLA~\cite{huang2026coworld} & Qwen3 & 99.1 & 97.0 & 99.6 &
        99.9 & 87.9 & 98.5 & 97.7 & 98.2 & 86.2 & 86.2 & 90.0 \\
        DreamerAD~\cite{yang2026dreamerad} & Transformer-1.3B & 98.0 & 97.2 &
        99.5 & 99.8 & 87.8 & 97.4 & 97.5 & 98.3 & 72.4 & -- & 87.7 \\
        Latent-WAM~\cite{wang2026latentwam} & DINOv2-B & 98.1 & 97.3 & 99.6 &
        99.8 & 87.7 & 97.3 & 97.6 & 98.1 & 87.3 & -- & 89.3 \\
        DriveFuture~\cite{hong2026drivefuture} & V2-99 & 98.8 & 99.1 & 99.6 &
        99.9 & 86.6 & 98.4 & 96.4 & 98.3 & 74.8 & 86.4 & 89.9 \\
        Discrete-WAM~\cite{yao2026discretewam} & Transformer-1B & 98.5 & 98.2 &
        99.7 & 99.8 & 90.5 & 97.9 & 97.2 & 98.3 & 78.1 & 87.0 & 90.4 \\
        \midrule
        \textbf{WA-JEPA (ours)} & ViT-L & 99.4 & 98.2 & 99.7 &
        99.9 & 87.8 & 98.9 & 98.3 & 98.3 & 88.1 &
        \textbf{88.0} & \textbf{91.7} \\
        \bottomrule
    \end{tabular}
    }
    % \caption{Comparison on NAVSIM-v2 \texttt{navtest}. Methods are grouped by
    % modeling paradigm. Backbone denotes the principal visual encoder or VLM
    % backbone reported by each method. $^\dagger$Uses perception annotations.
    % EPDMS$^{*}$ denotes scores obtained before the human-behavior filtering
    % fix, whereas EPDMS denotes corrected scores.}
\caption{Comparison with state-of-the-art methods on NAVSIM-v2
\texttt{navtest}. Methods are grouped by modeling paradigm.
Backbone lists the primary visual or vision-language backbone used
by each method. $^\dagger$ marks results using
auxiliary simulator-derived supervision. EPDMS$^*$ refers to scores
obtained before correction of the human-reference penalty-filter
aggregation, whereas EPDMS reports the corrected scores.}
    \label{tab:navsim_v2}
\end{table*}

% \begin{table}[t]
%     \centering
%     {\small
%     \setlength{\tabcolsep}{3pt}
%     \begin{tabular}{@{}lrrrrrr@{}}
%         \toprule
%         Method & NC$\uparrow$ & DAC$\uparrow$ & TTC$\uparrow$ &
%         Comf.$\uparrow$ & EP$\uparrow$ & PDMS$\uparrow$ \\
%         \midrule
%         \multicolumn{7}{@{}l}{\textbf{End-to-End Methods}} \\
%         TransFuser & 97.7 & 92.8 & 92.8 & 100 & 79.2 & 84.0 \\
%         DiffusionDrive & 98.2 & 96.2 & 94.7 & 100 & 82.2 & 88.1 \\
%         Drive-JEPA & 98.7 & 96.2 & 95.5 & 100 & 82.9 & 89.0 \\
%         \midrule
%         \multicolumn{7}{@{}l}{\textbf{VLA Methods}} \\
%         Uni-World VLA & 98.7 & 96.7 & 96.1 & 100 & 83.2 & 89.4 \\
%         CoWorld-VLA & 99.1 & 96.9 & 96.4 & 100 & 83.9 & 89.9 \\
%         DriveVLA-W0 & 98.7 & 99.1 & 95.3 & 99.3 & 83.3 & 90.2 \\
%         WAM-Flow & 99.2 & 98.3 & 97.0 & 99.7 & 82.3 & 90.3 \\
%         WAM-Diff & 99.1 & 98.3 & 96.5 & 99.9 & 84.4 & 91.0 \\
%         DriveWorld-VLA & 99.1 & 98.2 & 96.1 & 100 & 85.9 & 91.3 \\
%         \midrule
%         \multicolumn{7}{@{}l}{\textbf{World-Action and World-Model Methods}} \\
%         DriveLaW & 99.0 & 97.1 & 96.7 & 100 & 81.3 & 89.1 \\
%         DriveFuture & 98.8 & 99.1 & 95.4 & 100 & 84.2 & 90.7 \\
%         \midrule
%         \textbf{WA-JEPA (ours)} & 99.5 & 98.3 & 97.7 &
%         100 & 85.0 & \textbf{91.8} \\
%         \bottomrule
%     \end{tabular}
%     }
%     \caption{Comparison on NAVSIM-v1 \texttt{navtest}.}
%     \label{tab:navsim_v1}
% \end{table}
\section{Experiments}
\label{sec:experiment}

\subsection{Experimental Setup}

\paragraph{Datasets, benchmarks, and metrics.}
In Stage~1, the model is pre-trained on multi-view driving videos from nuPlan, and Stage~2
is trained on the official NAVSIM \texttt{navtrain} split. Evaluation is
performed on the held-out \texttt{navtest} split under NAVSIM-v1 and
NAVSIM-v2~\cite{dauner2024navsim,cao2025navsim}, using the official PDMS and
EPDMS scores and their sub-metrics, which are defined in Appendix~\ref{app:metrics}.

\paragraph{Implementation details.}
The model takes four historical frames from
the left, front, right, and rear cameras at $256\times512$ resolution and
predicts eight actions at 2\,Hz. The visual encoder is initialized from the
V-JEPA 2 ViT-L backbone pre-trained in Stage~1. The future scene and
action streams are jointly optimized with AdamW, bfloat16 precision, and
DeepSpeed ZeRO-2. Stage~1 and Stage~2 use 64 and 32 NVIDIA A800 GPUs,
respectively, with a per-GPU batch size of 4. The encoder, scene projector, and
joint predictor use learning rates of $1\times10^{-5}$,
$1\times10^{-4}$, and $1.5\times10^{-4}$, respectively, with weight decay
$0.04$. 
% At inference, the flow predictor uses 12 sampling steps with a shared set of random seeds, and the results are averaged across these seeds. 
At inference, the WA-JEPA Joint future-action predictor uses 4 sampling steps. 
We evaluate its stochastic predictions using a fixed set of 10 seeds
and report the mean, while deterministic baselines are evaluated once.
Detailed settings are provided in Appendix~\ref{app:additional-results}.

\paragraph{Baselines.}
We compare WA-JEPA with representative E2E, VLA, and WAM baselines
under compatible input modalities and evaluation protocols.

% These floats are intentionally declared before the results subsections so
% that the full-width PCA figure can occupy an experimental-page top rather
% than being deferred to the bibliography.
\begin{table}[!t]
\centering
{%
\small
\setlength{\tabcolsep}{3.4pt}
\begin{tabular}{@{}lrrrrr@{}}
\toprule
& WA-JEPA & LTF & DrivoR & UniAD & VAD \\
\midrule
\multicolumn{6}{@{}l}{\emph{All 436 scenarios}}\\
NC       & \textbf{0.6856} & 0.4428 & 0.5217 & 0.6555 & 0.4117 \\
DAC      & \textbf{0.9635} & 0.9275 & 0.9559 & 0.9320 & 0.9028 \\
TTC      & \textbf{0.6120} & 0.3751 & 0.4620 & 0.5156 & 0.2798 \\
Comf.    & 0.6620 & 0.9478 & 0.9390 & 0.6633 & \textbf{0.9534} \\
PDMS     & \textbf{0.5717} & 0.3653 & 0.4475 & 0.4940 & 0.2831 \\
RC       & \textbf{0.5689} & 0.3804 & 0.4721 & 0.4383 & 0.3006 \\
HD-Score & \textbf{0.4462} & 0.2310 & 0.3252 & 0.3124 & 0.1393 \\
\midrule
\multicolumn{6}{@{}l}{\emph{HD-Score by difficulty level}}\\
Easy ($n{=}80$)     & \textbf{0.7977} & 0.6608 & 0.7799 & 0.6395 & 0.4197 \\
Medium ($n{=}157$)  & \textbf{0.5563} & 0.1547 & 0.2911 & 0.3718 & 0.0849 \\
Hard ($n{=}96$)     & \textbf{0.3060} & 0.1204 & 0.2000 & 0.2099 & 0.0770 \\
Extreme ($n{=}103$) & 0.1362 & 0.1167 & \textbf{0.1407} & 0.0632 & 0.0626 \\
\bottomrule
\end{tabular}
}
\caption{Zero-shot closed-loop results on HUGSIM. Values are on the
$[0,1]$ scale; higher is better, and the best result in each row is bold.}
\label{tab:hugsim_main}
\end{table}

\begin{figure*}[!t]
  \centering
  \includegraphics[width=\textwidth]
  {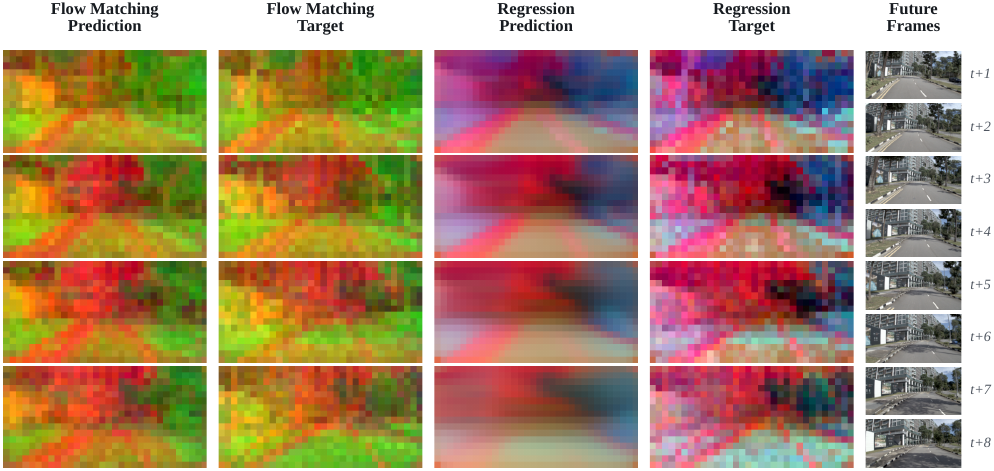}
  \caption{Target-referenced PCA visualization of future latent
  predictions from flow matching (FM) and direct regression (Reg.).
  For each method, a separate PCA basis is fitted to its EMA target
  representations and applied to both target and predicted latents.
  Each map represents two consecutive future frames.}
  \label{fig:fm_vs_reg}
\end{figure*}

% \subsection{Main Results}
\subsection{Comparison with Existing Methods}
\paragraph{Quantitative results.}
As shown in Table~\ref{tab:navsim_v2}, WA-JEPA achieved an EPDMS of 91.7 on
NAVSIM-v2, exceeding the best-performing E2E method, SparseDriveV2, by
1.6 and the best-performing WAM method, Discrete-WAM, by 1.3. On
NAVSIM-v1 (Table~\ref{tab:navsim_v1}), WA-JEPA attained a PDMS of 91.8.

\begin{table}[!t]
    \centering
    {\small
    \setlength{\tabcolsep}{3pt}
    \begin{tabular}{@{}lrrrrrr@{}}
        \toprule
        Method & NC$\uparrow$ & DAC$\uparrow$ & TTC$\uparrow$ &
        Comf.$\uparrow$ & EP$\uparrow$ & PDMS$\uparrow$ \\
        \midrule
        \multicolumn{7}{@{}l}{\textbf{End-to-End Methods}} \\
        TransFuser & 97.7 & 92.8 & 92.8 & 100 & 79.2 & 84.0 \\
        DiffusionDrive & 98.2 & 96.2 & 94.7 & 100 & 82.2 & 88.1 \\
        Drive-JEPA & 98.7 & 96.2 & 95.5 & 100 & 82.9 & 89.0 \\
        \midrule
        \multicolumn{7}{@{}l}{\textbf{VLA Methods}} \\
        ReCogDrive & 97.9 & 97.3 & 94.9 & 100 & 87.3 & 90.8 \\
        WAM-Flow & 99.2 & 98.3 & 97.0 & 99.7 & 82.3 & 90.3 \\
        WAM-Diff & 99.1 & 98.3 & 96.5 & 99.9 & 84.4 & 91.0 \\
        \midrule
        \multicolumn{7}{@{}l}{\textbf{World-Action and World-Model Methods}} \\
        CoWorld-VLA & 99.1 & 96.9 & 96.4 & 100 & 83.9 & 89.9 \\
        DriveVLA-W0 & 98.7 & 99.1 & 95.3 & 99.3 & 83.3 & 90.2 \\
        DriveWorld-VLA & 99.1 & 98.2 & 96.1 & 100 & 85.9 & 91.3 \\
        DriveLaW & 99.0 & 97.1 & 96.7 & 100 & 81.3 & 89.1 \\
        DriveFuture & 98.8 & 99.1 & 95.4 & 100 & 84.2 & 90.7 \\
        \midrule
        \textbf{WA-JEPA (ours)} & 99.5 & 98.3 & 97.7 &
        100 & 85.0 & \textbf{91.8} \\
        \bottomrule
    \end{tabular}
    }
    \caption{Comparison on NAVSIM-v1 \texttt{navtest}.}
    \label{tab:navsim_v1}
\end{table}

\paragraph{Zero-shot closed-loop generalization.}
We compare WA-JEPA with LTF~\cite{chitta2022transfuser},
DrivoR~\cite{kirby2026drivor}, UniAD~\cite{hu2023planning}, and
VAD~\cite{jiang2023vad} on 436 HUGSIM scenarios~\cite{zhou2024hugsim}.
WA-JEPA's two training stages use
neither HUGSIM-rendered observations nor any of its four source
datasets; DrivoR is similarly source-disjoint. All methods follow
the common protocol detailed in Appendix~\ref{app:hugsim}, sharing
the same scenarios, controller, commands, aggregation, and metric
implementation while retaining their native sensor configurations.

As shown in Table~\ref{tab:hugsim_main}, WA-JEPA achieves the best NC,
DAC, TTC, PDMS, RC, and HD-Score, improving HD-Score from $0.3252$ to
$0.4462$. This result demonstrates that joint world--action
pre-training transfers effectively to closed-loop control, with the
largest gains on the medium and hard levels.

\begin{figure}[!t]
  \centering
  \includegraphics[width=\linewidth]
  {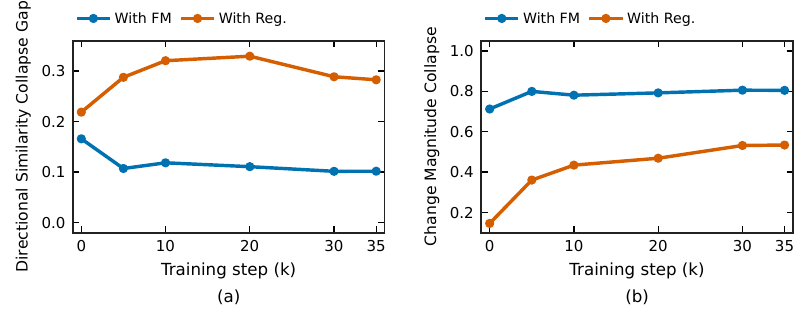}
  \caption{Temporal representation preservation with flow matching
  (FM) and direct regression (Reg.). A lower directional-similarity
  collapse gap and a change-magnitude collapse closer to 1 indicate
  better preservation of the target temporal dynamics.}
  \label{fig:gap-delta-comparison}
\end{figure}

% Figure~\ref{fig:hugsim_vis} visualizes closed-loop rollouts of WA-JEPA. Across
% turning, oncoming-vehicle encounters, and overtaking, the plans follow the
% drivable corridor and keep lateral clearance from the reactive agent on all four
% source datasets, despite the appearance gap between the rendered observations
% and the training data.

\paragraph{Qualitative evaluation.}
Qualitative results are provided in
Appendix~\ref{app:qualitative-results}, including closed-loop HUGSIM
rollouts and NAVSIM trajectory comparisons.

% \paragraph{PCA visualization.}
% Figure~\ref{fig:fm_vs_reg} provides a complementary PCA
% visualization over the eight-frame prediction horizon. Flow
% matching retains spatial structures aligned with the teacher,
% including the right-side lane markings at the farthest prediction
% step, whereas regression progressively loses local contrast.
% This observation is consistent with the quantitative results.
\subsection{Ablation Studies}
\paragraph{Vision encoder initialization.}
Table~\ref{tab:ablations}(a) isolates the contribution of the visual
encoder. Every variant is trained directly in Stage~2 without any
Stage~1 pre-training, so the models share the same Stage~2
architecture, training data, and optimization protocol and differ only
in how the encoder is initialized. The publicly released V-JEPA 2
weights reach an EPDMS of 89.5, exceeding the strongest alternatives,
MAE and DINOv3, by 5.7 EPDMS, whereas the three image-level
self-supervised and vision--language initializations lie within 0.7
EPDMS of one another. The gap therefore tracks the V-JEPA 2
pre-training objective rather than the choice among image-level
objectives, which motivates adopting a V-JEPA 2 encoder as the
backbone of our world-action model.

\begin{table}[!htbp]
    \centering

    % =========================================================
    % First row: Panel (a) + Panel (b)
    % =========================================================
    \noindent
    % -------------------- Panel (a) --------------------
    \begin{minipage}[t]{0.39\columnwidth}
        \vspace{0pt}
        \centering
        {\footnotesize
        \setlength{\tabcolsep}{2.5pt}
        \renewcommand{\arraystretch}{1.0}
    
        \begin{tabular}[t]{@{}lc@{}}
            \toprule
            Encoder & EPDMS$\uparrow$ \\
            \midrule
    
            MAE
                & 83.8 \\
    
            SigLIP2
                & 83.1 \\
    
            DINOv3
                & 83.8 \\
    
            V-JEPA 2
                & \textbf{89.5} \\
    
            \bottomrule
        \end{tabular}
    
        \vspace{2pt}
    
        \textbf{(a) Encoder}
        }
    \end{minipage}
    \hfill
    % -------------------- Panel (b) --------------------
    \begin{minipage}[t]{0.59\columnwidth}
        \vspace{0pt}
        \centering
        {\footnotesize
        \setlength{\tabcolsep}{3.5pt}
        \renewcommand{\arraystretch}{1.0}
    
        % Fix the widths of the two mask columns
        \def\MaskCell#1{\makebox[4.8em][c]{#1}}
    
        \begin{tabular}[t]{@{}ccc@{}}
            \toprule
    
            \MaskCell{Patch-mask}
                & \MaskCell{Full-mask}
                & EPDMS$\uparrow$ \\
    
            \midrule
    
                       &            & 89.5 \\
            \checkmark &            & 91.0 \\
                       & \checkmark & 91.3 \\
            \checkmark & \checkmark & \textbf{91.7} \\
    
            \bottomrule
        \end{tabular}
    
        \vspace{2pt}
    
        \textbf{(b) Stage~1 Pre-training}
        }
    \end{minipage}
    
    % Explicitly end the first row
    \par\vspace{7pt}
    
    % =========================================================
    % Second row: Panel (c), centered
    % =========================================================
    \noindent\hfill
    % -------------------- Panel (c) --------------------
    \begin{minipage}[t]{0.72\columnwidth}
        \vspace{0pt}
        \centering
        {\footnotesize
        \setlength{\tabcolsep}{3pt}
        \renewcommand{\arraystretch}{1.0}
    
        % Equal widths for FM and Reg.
        \def\PredCell#1{\makebox[2.8em][c]{#1}}
    
        \begin{tabular}[t]{@{}cccc@{}}
            \toprule
    
            \multirow{2}{*}{Joint}
                & \multicolumn{2}{c}{Future Pred.}
                & \multirow{2}{*}{EPDMS$\uparrow$} \\
    
            \cmidrule(lr){2-3}
    
                & \PredCell{FM}
                & \PredCell{Reg.}
                & \\
    
            \midrule
    
                       &            &            & 89.9 \\
                       & \checkmark &            & 90.8 \\
            \checkmark &            &            & 91.1 \\
            \checkmark &            & \checkmark & 90.7 \\
            \checkmark & \checkmark &            & \textbf{91.7} \\
    
            \bottomrule
        \end{tabular}
    
        \vspace{2pt}
    
        \textbf{(c) Stage~2 Component}
        }
    \end{minipage}
    \hfill\mbox{}

    \caption{Ablations on NAVSIM-v2 \texttt{navtest}.
    (a) Vision encoder selection.
   (b) Masking strategy in Stage~1. The first row represents the baseline that skips Stage~1 pre-training and directly uses the original pre-trained V-JEPA 2 for Stage~2 training.
    (c) Joint future--action predictor and future-prediction
    methods in Stage~2. FM and Reg. denote flow matching and
    direct regression, respectively. The first row denotes an action-only Stage~2 baseline initialized from the Stage~1 checkpoint and fine-tuned with aciton supervision, without joint scene-action modeling or future prediction.}

    \label{tab:ablations}
\end{table}

\paragraph{Stage~1 masking strategies.}
Table~\ref{tab:ablations}(b) ablates the masking design in Stage~1.
The first row directly uses the original pretrained V-JEPA~2
checkpoint without our Stage~1 pre-training on nuPlan, serving as
the no-Stage~1 baseline with an EPDMS of 89.5. Applying Patch-mask
alone improves EPDMS to 91.0 by providing partial future context for
representation learning, while Full-mask achieves 91.3 by enforcing
strictly causal past-to-future prediction. Combining both strategies
yields the best performance of 91.7, outperforming the individual
variants by 0.7 and 0.4, respectively. This demonstrates that
Patch-mask and Full-mask provide complementary training signals.

\paragraph{Stage~2 scene--action coupling and future prediction.}
Table~\ref{tab:ablations}(c) ablates scene--action modeling and
future-latent supervision in Stage~2. The first cascaded baseline
uses only historical latent features through cross-attention and
obtains 89.9 EPDMS. Adding a separate flow-based future predictor
and injecting its predicted future latents through the same
cross-attention design improves EPDMS to 90.8. Under joint modeling,
the baseline without explicit future-latent supervision achieves
91.1. Adding direct regression reduces EPDMS to 90.7, whereas
flow-based future prediction yields the best result of 91.7.
These results show that future-scene prediction and joint
scene--action modeling are complementary, while the choice of
prediction objective also matters.

\paragraph{Future representation analysis.}
To examine how the two prediction objectives affect future
representations, Fig.~\ref{fig:gap-delta-comparison} reports two
target-referenced metrics. The directional similarity collapse gap
measures excessive cross-step similarity relative to the target
representations (lower is better), while the change-magnitude
collapse measures the predicted temporal variation relative to the
targets (closer to 1 is better). Detailed definitions are provided in Appendix~\ref{app:temporal-metrics}. Compared with direct
regression, flow matching reduces the directional similarity collapse
gap from \(0.30\) to \(0.10\) and increases the change-magnitude
collapse from \(0.45\) to \(0.80\). Figure~\ref{fig:fm_vs_reg}
provides complementary qualitative evidence: flow matching retains
clearer spatial structures over the prediction horizon, whereas
direct regression becomes progressively smoother.

\section{Conclusion}
We presented WA-JEPA, a V-JEPA-native world-action model that extends
V-JEPA~2 from visual representation learning to future-predictive
planning for autonomous driving. By combining hybrid future-masked
pre-training, prediction of future latents via flow matching, and
joint future--action modeling, WA-JEPA learns future scene dynamics
and actions within a shared spatiotemporal latent space. WA-JEPA
achieves 91.7 EPDMS on NAVSIM-v2 \texttt{navtest} and a 0.4462
HD-Score on 436 HUGSIM closed-loop scenarios, demonstrating strong
open-loop planning and zero-shot closed-loop generalization.
% Extending this evaluation to real-world closed-loop driving remains
% future work.

% \section*{Acknowledgments}

\clearpage

\appsection{HUGSIM Closed-Loop Evaluation}
\label{app:hugsim}

% \paragraph{Protocol.}
% The 436 scenarios span four source datasets and four difficulty levels; the difficulty distribution is reported in Table~\ref{tab:hugsim_main} of the main text and the dataset distribution in Table~\ref{tab:hugsim-dataset}. All methods use the same scenarios, heading-corrected controller, ground-truth driving commands, and aggregation procedure. Sensor configurations follow the original methods: LTF uses three front-view cameras, while the others use four.

% We rescore all evaluated methods using the corrected HUGSIM evaluation code released by the authors of \emph{Driving on Registers} (DrivoR), ensuring consistent metric computation across methods. DrivoR was originally evaluated on an earlier HUGSIM release containing 345 scenarios, whereas our evaluation uses the current 436-scenario release. Its published scores are therefore not directly comparable to those reported here. For DrivoR, we report the variant trained on the combined NAVSIM training and validation sets without the additional simulated training data introduced in SimScale. In contrast, Stage 2 of WA-JEPA uses only the NAVSIM training set.

\paragraph{Protocol.}
The 436 scenarios span four source datasets and four difficulty
levels; the difficulty distribution is reported in
Table~\ref{tab:hugsim_main} of the main text and the dataset
distribution in Table~\ref{tab:hugsim-dataset}. All methods use the
same scenarios, ground-truth driving commands, aggregation procedure,
and HUGSIM controller and evaluator. Specifically, we use HUGSIM at
commit
\href{https://github.com/hyzhou404/HUGSIM/commit/ead17f2ad97f71fd21fa6f66237a7c05364ed98e}
{\texttt{ead17f2}}, which applies the trajectory-to-heading
coordinate-order correction introduced in
\href{https://github.com/hyzhou404/HUGSIM/pull/57}{PR~\#57}.
Sensor configurations follow the original methods: LTF uses three
front-view cameras, while the others use four.

We rescore all evaluated methods using this fixed code snapshot,
ensuring consistent controller and metric implementations. DrivoR was
originally evaluated on an earlier HUGSIM release containing 345
scenarios, whereas our evaluation uses the current 436-scenario set.
Its published scores are therefore not directly comparable to those
reported here. For DrivoR, we report the variant trained on the
combined NAVSIM training and validation sets without the additional
simulated training data introduced in SimScale. In contrast, Stage~2
of WA-JEPA uses only the NAVSIM training set.

\paragraph{Aggregation robustness.}
Within each difficulty level we average scenarios uniformly inside a source
dataset and then average the four datasets uniformly, as in the HUGSIM
benchmark. The single overall HD-Score is obtained by weighting the four
difficulty levels by their scenario counts ($80/157/96/103$), following
subsequent work on this benchmark. Table~\ref{tab:hugsim-aggregation} reports
two alternative rules, uniform averaging over the four datasets and uniform
averaging over all 436 scenarios. The method ranking in the main text is
unchanged under both.

% \paragraph{Per-dataset results.}
% Table~\ref{tab:hugsim-dataset} breaks the HD-Score down by source dataset. WA-JEPA obtains the highest HD-Score on all four datasets, including the three that none of its training stages has seen.
\paragraph{Per-dataset results.}
Table~\ref{tab:hugsim-dataset} reports the HD-Score separately for
each HUGSIM source dataset. None of the four datasets is used for
either Stage~1 or Stage~2 training. WA-JEPA achieves the highest
HD-Score on every dataset, indicating consistent source-disjoint
generalization across diverse visual domains.

\vspace{4pt}

\begin{center}
{%
\scriptsize
\setlength{\tabcolsep}{2.4pt}
\begin{tabular}{@{}lrrrrr@{}}
\hline
Aggregation & WA-JEPA & LTF & DrivoR & UniAD & VAD \\
\hline
Primary             & \textbf{0.4462} & 0.2310 & 0.3252 & 0.3124 & 0.1393 \\
Dataset-uniform     & \textbf{0.4483} & 0.2300 & 0.3246 & 0.3085 & 0.1304 \\
Scenario-uniform    & \textbf{0.4464} & 0.2243 & 0.3194 & 0.3082 & 0.1266 \\
\hline
\end{tabular}
}
\captionof{table}{HD-Score under three aggregation rules. \emph{Primary} is the
rule described above; the other two rows average uniformly over
the four datasets and over all 436 scenarios, respectively. The best result in
each row is bold.}
\label{tab:hugsim-aggregation}
\end{center}

\begin{center}
{%
\scriptsize
\setlength{\tabcolsep}{2.2pt}
\begin{tabular}{@{}lrrrrrr@{}}
\hline
Dataset & \(n\) & WA-JEPA & LTF & DrivoR & UniAD & VAD \\
\hline
nuScenes  & 88  & \textbf{0.4725} & 0.3334 & 0.3830 & 0.3405 & 0.2069 \\
KITTI-360 & 113 & \textbf{0.2963} & 0.0969 & 0.2175 & 0.0550 & 0.0272 \\
Waymo     & 108 & \textbf{0.5542} & 0.2478 & 0.4025 & 0.4372 & 0.1376 \\
PandaSet  & 127 & \textbf{0.4702} & 0.2419 & 0.2955 & 0.4012 & 0.1500 \\
\hline
\end{tabular}
}
% \captionof{table}{HD-Score by source dataset. The best result in each row is
% bold.}
\captionof{table}{Per-dataset HD-Scores on the 436 HUGSIM scenarios.
$n$ denotes the number of scenarios. The best result in each row is
bold.}
\label{tab:hugsim-dataset}
\end{center}

\appsection{Qualitative Results}
\label{app:qualitative-results}

\begin{figure*}[t]
  \centering
  \includegraphics[width=\textwidth]{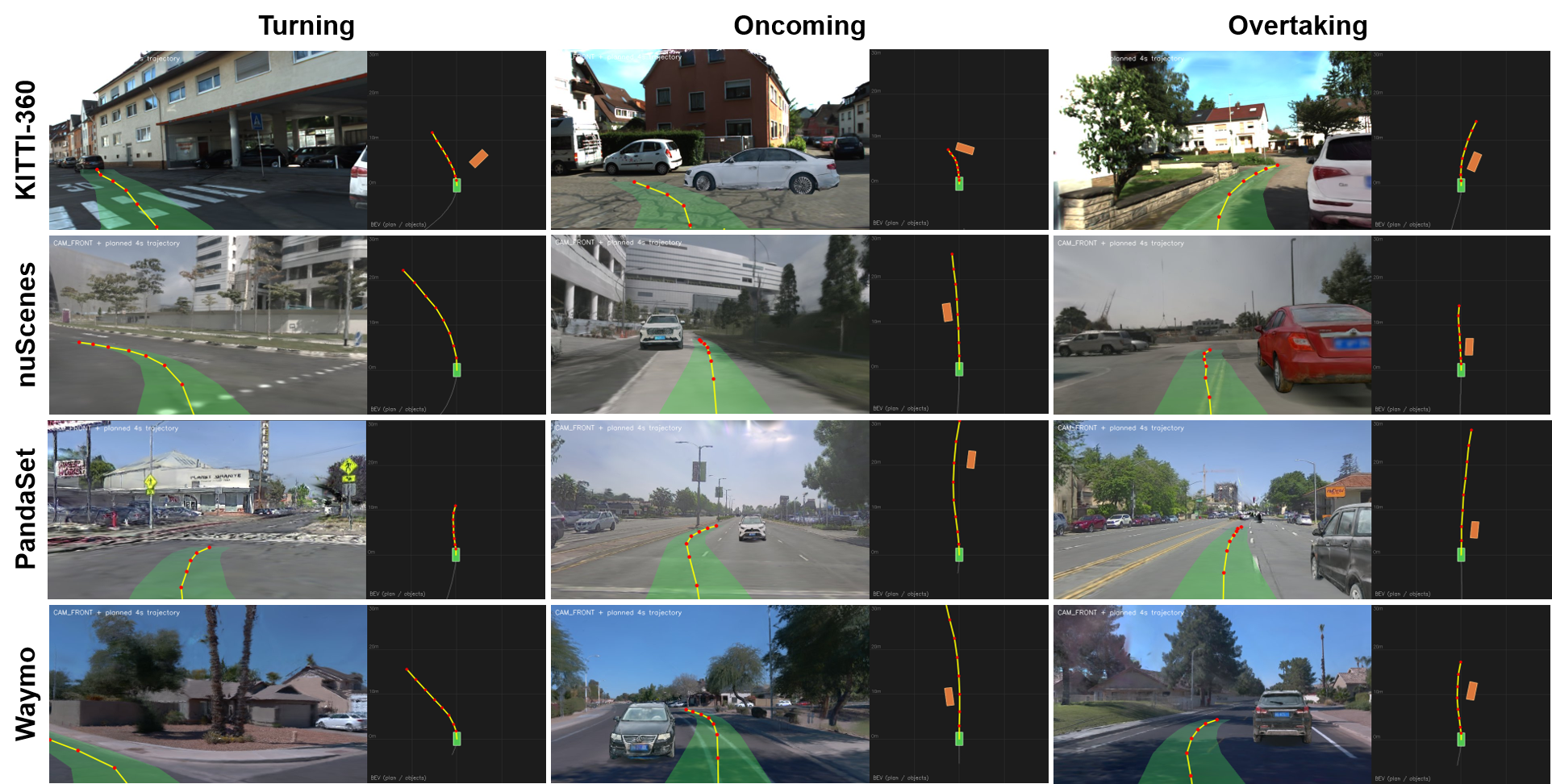}
  \caption{Zero-shot closed-loop rollouts of WA-JEPA on HUGSIM. Rows
  correspond to the four source datasets and columns to three scenario types.
  Each pair shows the front camera with the projected 4\,s plan (left) and the
  BEV view with the planned trajectory and detected objects (right). The green
  region marks the drivable corridor, the yellow--red curve the planned
  trajectory, the green box the ego vehicle, and orange boxes other agents.}
  \label{fig:hugsim_vis}
\end{figure*}

\paragraph{HUGSIM closed-loop rollouts.}
Figure~\ref{fig:hugsim_vis} visualizes zero-shot closed-loop rollouts
across turning, oncoming-vehicle encounters, and overtaking. WA-JEPA
follows the drivable corridor and maintains lateral clearance from
reactive agents across all four source datasets despite the visual
domain gap between HUGSIM renderings and the training data.

\begin{figure*}[t]
  \centering
  \includegraphics[width=\textwidth]{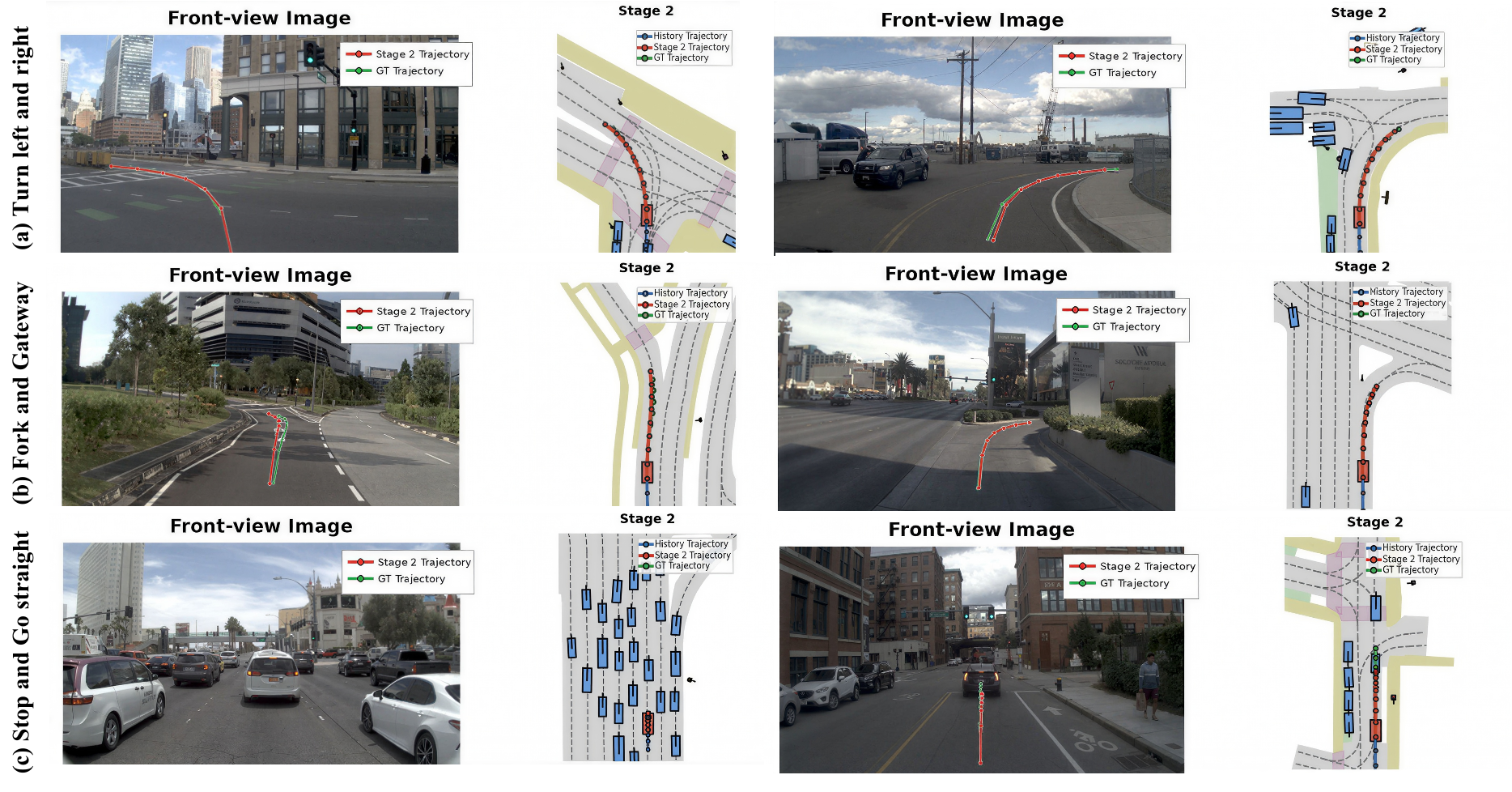}
  \caption{Trajectory predictions on representative NAVSIM scenarios:
  (a) left and right turns, (b) fork and gateway navigation, and
  (c) stopping and straight driving.}
  \label{fig:qualitative_traj_comparison}
\end{figure*}

\paragraph{NAVSIM trajectory predictions.}
Figure~\ref{fig:qualitative_traj_comparison} shows representative
Stage~2 predictions on NAVSIM. Across turning, fork and gateway
navigation, stopping, and straight-driving scenarios, the predicted
trajectories agree with the references in maneuver direction and
overall geometry, with only minor local deviations.

\appsection{Additional Experimental Details}
\label{app:additional-results}

\paragraph{Details of stop-gradient design.}
Although the scene output stream conditions on the action tokens during the
forward pass, gradients from the scene prediction loss are blocked at the
action-token interface and therefore do not update the action stream. In
contrast, the action output stream attends to differentiable historical
context and future scene tokens, allowing action supervision to shape the
learned scene representations through the joint interaction module. This
asymmetric gradient design preserves future scene prediction while encouraging
the scene representations to capture future dynamics that are relevant to ego
planning.

\paragraph{Inference details and seed-level variability.}

The WA-JEPA flow predictor uses 4 sampling steps. For all experiments involving stochastic noise initialization, we use a fixed set of ten seeds. For each seed, we reinitialize the sampling noise and run the full evaluation while keeping the model parameters, scenarios, and inference settings unchanged. All sub-metrics and EPDMS are computed independently for each seed, with EPDMS following Eq.~\eqref{eq:epdms}. We report their arithmetic mean before rounding.

\begin{table}[t]
\centering
{%
\small
\begin{tabular}{lc}
\toprule
Statistic & EPDMS \\
\midrule
Number of seeds & 10 \\
Mean & 91.7014 \\
Standard deviation & 0.0531 \\
Standard error & 0.0168 \\
95\% $t$-confidence interval & $[91.6634,\,91.7393]$ \\
Median & 91.6960 \\
Range & $[91.6294,\,91.8070]$ \\
\bottomrule
\end{tabular}
}
\caption{Seed-level EPDMS variability for the main WA-JEPA experiment over a fixed set of ten seeds.}
\label{tab:seed_variability}
\end{table}

As shown in Table~\ref{tab:seed_variability}, the main WA-JEPA experiment achieves a mean EPDMS of $91.7014$, which is reported as $91.7$ in the main text after rounding. Methods without stochastic noise initialization are evaluated once, while all reproduced baselines use their native inference procedures.

\appsection{NAVSIM Evaluation Metrics}
\label{app:metrics}

Evaluation follows the official NAVSIM protocol, with all scores computed by
the pseudo-simulator on \texttt{navtest}. This section defines each
sub-metric and the aggregation rules used in
Tables~\ref{tab:navsim_v2} and~\ref{tab:navsim_v1} of the main text.

\paragraph{Sub-metrics.}
NAVSIM-v1 uses five sub-metrics:
\emph{no-at-fault collision} (NC), which is $0$ if the ego vehicle causes a
collision and $1$ otherwise (collisions for which the ego is not responsible,
e.g.\ being rear-ended, are excluded);
\emph{drivable-area compliance} (DAC), which is $0$ if any part of the ego
footprint leaves the drivable area and $1$ otherwise;
\emph{time-to-collision} (TTC), which is $1$ when the minimum time-to-collision
along the rollout stays above a safety threshold under a constant-velocity
projection of the ego and surrounding agents;
\emph{comfort} (Comf.), which is $1$ when longitudinal and lateral
accelerations, jerk, and yaw rate remain within the human-driving bounds
estimated from the dataset; and
\emph{ego progress} (EP), the ratio of the ego's traveled distance along the
route centerline to that of the privileged PDM-Closed planner, clipped
to $[0,1]$.

NAVSIM-v2 additionally introduces:
\emph{driving-direction compliance} (DDC), which penalizes driving against the
nominal direction of the current lane, with a graded penalty depending on the
magnitude of the violation;
\emph{traffic-light compliance} (TLC), which is $0$ if the ego crosses a
stop line under a red light and $1$ otherwise;
\emph{lane keeping} (LK), which measures whether the ego stays within its
assigned lane corridor over the horizon;
\emph{history comfort} (HC), which evaluates comfort bounds jointly over the
past trajectory and the planned trajectory, thereby penalizing abrupt
transitions at the planning boundary; and
\emph{extended comfort} (EC), which measures the consistency of the
kinematic profile across the two stages of the pseudo-simulation, so that
trajectories that change sharply between rollouts are penalized.

\paragraph{Aggregation.}
Sub-metrics are grouped into multiplicative penalties $\mathcal{M}_{\text{pen}}$
and a weighted average $\mathcal{M}_{\text{avg}}$. The NAVSIM-v1 Predictive
Driver Model Score (PDMS) is
\begin{equation}
\label{eq:pdms}
\text{PDMS} =
\underbrace{\text{NC}\cdot\text{DAC}}_{\mathcal{M}_{\text{pen}}}
\cdot
\frac{5\,\text{TTC} + 2\,\text{Comf.} + 5\,\text{EP}}{12},
\end{equation}
so that a single safety violation drives the score to zero, while the remaining
terms trade off safety margin, comfort, and progress.

The NAVSIM-v2 Extended PDMS (EPDMS) additionally applies a
human-reference penalty filter before aggregating the sub-metrics.
Let $s_{i,m}^{\mathrm{agent}}$ and
$s_{i,m}^{\mathrm{human}}$ denote the agent and human-reference
scores, respectively, for metric $m$ in scenario $i$. For original
first-stage pseudo-simulation scenarios, the corrected sub-metric is
defined as
\begin{equation}
\label{eq:human-filter}
\tilde{s}_{i,m}
=
\operatorname{filter}_{m}
\left(
s_{i,m}^{\mathrm{agent}},
s_{i,m}^{\mathrm{human}}
\right)
=
\begin{cases}
1,
&
\substack{
m\in\mathcal{M}_{\mathrm{filt}},\\
s_{i,m}^{\mathrm{human}}=0,
}
\\
s_{i,m}^{\mathrm{agent}},
& \text{otherwise},
\end{cases}
\end{equation}
where
$\mathcal{M}_{\mathrm{filt}}
=\{\mathrm{NC},\mathrm{DAC},\mathrm{DDC},\mathrm{TLC},
\mathrm{EP},\mathrm{TTC},\mathrm{LK},\mathrm{HC}\}$.
Synthetic second-stage pseudo-simulation scenarios do not use this
filter, and EC is computed subsequently from the paired rollouts.
Thus, the filter does not remove scenarios; it suppresses a
metric-specific penalty when the human reference incurs the same
violation.

Using the filtered sub-metrics, the per-scenario extended score is
\begin{equation}
\label{eq:epdms-scenario}
\begin{aligned}
q_i
={}&
\tilde{s}_{i,\mathrm{NC}}
\tilde{s}_{i,\mathrm{DAC}}
\tilde{s}_{i,\mathrm{DDC}}
\tilde{s}_{i,\mathrm{TLC}}
\\
&\times
\frac{
5\tilde{s}_{i,\mathrm{EP}}
+5\tilde{s}_{i,\mathrm{TTC}}
+2\tilde{s}_{i,\mathrm{LK}}
+2\tilde{s}_{i,\mathrm{HC}}
+2\tilde{s}_{i,\mathrm{EC}}
}{16}.
\end{aligned}
\end{equation}

The final benchmark EPDMS follows the official two-stage NAVSIM-v2
aggregation. Let $\mathcal{G}$ denote the official mapping groups,
$\mathcal{I}_{g,b}^{(r)}$ the scenarios assigned to branch
$b\in\{1,2\}$ at pseudo-simulation stage $r\in\{1,2\}$, and
$\alpha_i$ the provided scene weight. Define
\begin{equation}
\bar q_{g,b}^{(r)}
=
\frac{
\sum_{i\in\mathcal{I}_{g,b}^{(r)}}\alpha_i q_i
}{
\sum_{i\in\mathcal{I}_{g,b}^{(r)}}\alpha_i
}.
\end{equation}
The reported score is
\begin{equation}
\label{eq:epdms}
\mathrm{EPDMS}
=
\frac{100}{|\mathcal{G}|}
\sum_{g\in\mathcal{G}}
\frac{1}{2}
\sum_{b=1}^{2}
\bar q_{g,b}^{(1)}\bar q_{g,b}^{(2)}.
\end{equation}

\paragraph{EPDMS$^{*}$ versus EPDMS.}
All corrected NAVSIM-v2 results are computed using the official
NAVSIM devkit at commit
\href{https://github.com/autonomousvision/navsim/commit/359c7f72304bfa8273e754224a213d3751bd2340}
{\texttt{359c7f7}}, which recomputes the multiplicative and weighted
terms after applying the human-reference filter. EPDMS$^{*}$ denotes
scores obtained with the pre-fix evaluator, whereas EPDMS denotes
scores obtained with this corrected protocol. The two settings are
therefore reported separately and compared only within the same
evaluation protocol.

\appsection{Temporal Representation Metrics}
\label{app:temporal-metrics}

We evaluate whether predicted future scene tokens preserve the temporal
variation of the corresponding EMA target tokens. Consistent with the main
paper, the comparison uses two target-referenced metrics: the
\emph{directional similarity collapse gap} (lower is better) and the
\emph{change-magnitude collapse} (closer to \(1\) is better). Both are
computed in the projected scene-token space.

\paragraph{Dynamic-token selection.}
Let \(\widehat{\mathbf z}_{i,r,t}\) and \(\mathbf z_{i,r,t}\) denote the
predicted and target features for prediction instance \(i\), camera--spatial
location \(r\), and future token step \(t\). To prevent static regions from
dominating the statistics, we rank locations by the target's mean
adjacent-step change,
\begin{equation}
\begin{aligned}
s_{i,r}
&=\frac{1}{F-1}\sum_{t=0}^{F-2}
\left\|\mathbf z_{i,r,t+1}-\mathbf z_{i,r,t}\right\|_2,\\
\mathcal A_i&=\operatorname{TopK}_{r}(s_{i,r}).
\end{aligned}
\label{eq:dynamic-token-selection}
\end{equation}
The same target-selected set \(\mathcal A_i\) is used for both methods and
both metrics. Selection is performed independently for each prediction
instance.

\paragraph{Directional similarity collapse gap.}
For a feature sequence \(\mathbf q\), define the mean cosine similarity over
all ordered pairs of distinct future steps at the selected locations as
\begin{equation}
\begin{aligned}
C(\mathbf q)
&=\frac{1}{N_{\mathrm{cos}}}
\sum_i\sum_{r\in\mathcal A_i}\sum_{t\ne u}
\operatorname{cos}
\left(\mathbf q_{i,r,t},\mathbf q_{i,r,u}\right),\\
\operatorname{cos}(\mathbf a,\mathbf b)
&=\frac{\mathbf a^{\mathsf T}\mathbf b}
{\max(\|\mathbf a\|_2,\epsilon)\max(\|\mathbf b\|_2,\epsilon)},
\quad \epsilon=10^{-6}.
\end{aligned}
\end{equation}
where \(N_{\mathrm{cos}}\) is the number of valid terms. The reported metric is
\begin{equation}
\Delta_{\mathrm{dir}}
=C(\widehat{\mathbf z})-C(\mathbf z).
\label{eq:directional-similarity-collapse-gap}
\end{equation}
A positive value indicates that predicted future steps are more mutually
similar than the targets and therefore exhibit additional directional
collapse. Accordingly, lower values indicate better target-relative temporal
preservation, as stated in the main text.

\paragraph{Change-magnitude collapse.}
We also compare the mean adjacent-step feature change,
\begin{equation}
\begin{aligned}
D(\mathbf q)
&=\frac{1}{N_{\Delta}}
\sum_i\sum_{r\in\mathcal A_i}\sum_{t=0}^{F-2}
\left\|\mathbf q_{i,r,t+1}-\mathbf q_{i,r,t}\right\|_2,\\
R_{\Delta}
&=\frac{D(\widehat{\mathbf z})}
{\max(D(\mathbf z),\epsilon)}.
\end{aligned}
\label{eq:change-magnitude-collapse}
\end{equation}
Although referred to as change-magnitude collapse in the main text,
\(R_{\Delta}\) is a ratio: \(R_{\Delta}=1\) means that the prediction
preserves the target's average temporal change, while values below \(1\)
indicate under-variation. Thus, values closer to \(1\) are better for the
comparisons reported in the main text.

\paragraph{Evaluation protocol.}
Both objectives use the same projected feature space, target-selected
locations, and global averaging over all valid prediction instances and token
pairs. For flow matching, the metric is computed from the one-step
\(x\)-prediction at the sampled training flow time, without running the
multi-step inference sampler; regression is evaluated from its direct latent
prediction. The summary values in the main text are arithmetic means of the
raw logged metrics over the common \(0\)--\(36\)k training interval. For the
standard setting, eight future frames with tubelet size \(2\) give \(F=4\)
future token steps, and \(K=64\) target-dynamic locations are selected from
the four-camera scene-token grid.

\end{document}